\documentclass[journal,twoside,print]{ieeecolor}

\usepackage{cite}
\usepackage{amsmath,amssymb,amsfonts}

\usepackage{graphicx}
\makeatletter
\def\endfigure{\end@float}
\def\endtable{\end@float}
\makeatother
\usepackage[caption=false,font=footnotesize]{subfig}
\usepackage{textcomp}
\usepackage{color}
\definecolor{subsectioncolor}{RGB}{0,0,0}

\usepackage{array}
\usepackage{placeins}
\usepackage{booktabs}
\usepackage{multicol}
\usepackage{multirow}
\usepackage{longtable}
\usepackage{indentfirst}
\usepackage{hyperref}
\usepackage[capitalise]{cleveref} 
\usepackage{float}
\usepackage{balance}
\usepackage{makecell}
\usepackage{rotating}
\usepackage{verbatim}
\usepackage{algorithm}
\usepackage{algpseudocode}

\def\BibTeX{{\rm B\kern-.05em{\sc i\kern-.025em b}\kern-.08em
T\kern-.1667em\lower.7ex\hbox{E}\kern-.125emX}}

\begin{document}
\title{Alzheimer's Disease Progression Prediction Based on Manifold Mapping of Irregularly Sampled Longitudinal Data}
\author{Xin Hong, Ying Shi, Yinhao Li, and Yen-Wei Chen
\thanks{This work is supported in part by the Scientific Research Start-up Fund Project for High-level Researchers of Huaqiao University under Grant 22BS105;in part by the Natural Science Foundation of Fujian Province, China under Grant 2022J01318. }
\thanks{Xin Hong is with the College of Computer Science and Technology, Huaqiao University, Xiamen 361021, China. and also with the Key Laboratory of Computer Vision and Machine Learning in Fujian Province, Xiamen 361021, China.(Corresponding author,email: xinhong@hqu.edu.cn; 10409035@qq.com). }
\thanks{Ying Shi is with the College of Computer Science and Technology, Huaqiao University, Xiamen 361021, China(email:23014083048@stu.hqu.edu.cn).}
\thanks{Yinhao Li is with the college of Information Science and Engineering,Ritsumeikan University, Ibaraki 525-8577, Japan. (email: yin-li@fc.ritsumei.ac.jp).}
\thanks{Yen-Wei Chen is with the college of Information Science and Engineering,Ritsumeikan University, Ibaraki 525-8577, Japan. (email: chen@is.ritsumei.ac.jp).}}

\maketitle

\begin{abstract}

The uncertainty of clinical examinations frequently leads to irregular observation intervals in longitudinal imaging data, posing challenges for modeling disease progression.
Most existing imaging-based disease prediction models operate in Euclidean space, which assumes a flat representation of data and fails to fully capture the intrinsic continuity and nonlinear geometric structure of irregularly sampled longitudinal images.
To address the challenge of modeling Alzheimer’s disease (AD) progression from irregularly sampled longitudinal structural Magnetic Resonance Imaging (sMRI) data, we propose a Riemannian manifold mapping, a Time-aware manifold Neural ordinary differential equation, and an Attention-based riemannian Gated recurrent unit (R-TNAG) framework. Our approach first projects features extracted from high-dimensional sMRI into a manifold space to preserve the intrinsic geometry of disease progression. On this representation, a time-aware Neural Ordinary Differential Equation (TNODE) models the continuous evolution of latent states between observations, while an Attention-based Riemannian Gated Recurrent Unit (ARGRU) adaptively integrates historical and current information to handle irregular intervals. This joint design improves temporal consistency and yields robust AD trajectory prediction under irregular sampling.
Experimental results demonstrate that the proposed method consistently outperforms state-of-the-art models in both disease status prediction and cognitive score regression. Ablation studies verify the contributions of each module, highlighting their complementary roles in enhancing predictive accuracy. Moreover, the model exhibits stable performance across varying sequence lengths and missing data rates, indicating strong temporal generalizability. Cross-dataset validation further confirms its robustness and applicability in diverse clinical settings.

\end{abstract}


\section{Introduction}
\label{sec:introduction}

\IEEEPARstart{A}{lzheimer's} disease (AD) is a neurodegenerative disorder characterized by progressive cognitive decline, where early detection can delay progression \cite{name1}. Owing to its non-invasive and cost-effective nature, structural Magnetic Resonance Imaging (sMRI) is widely used in clinical practice \cite{alzheimer20192019}. Longitudinal sMRI scans capture brain changes over time \cite{scahill2003longitudinal,fujita2023characterization}, and recent studies have employed deep learning to jointly model imaging and cognitive scores for improved trajectory prediction \cite{yoon2018estimating,nguyen2020predicting,jung2021deep}. However, clinical visits are typically irregular, producing unevenly sampled data that complicate temporal modeling and degrade prediction performance \cite{che2018recurrent}.

\begin{figure}[!t]
\centerline{\includegraphics[width=\columnwidth]{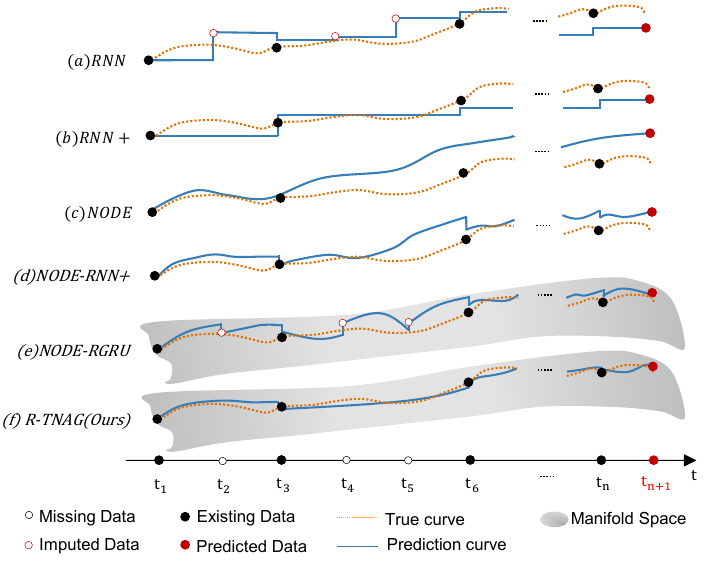}}
\caption{The figure presents a comparison between the predicted curves and the true curves for different models in addressing the irregular interval prediction problem.
(a) RNN are limited by the assumption of equal time intervals. When data are interpolated, errors can be easily introduced, and the prediction trajectory exhibits a stepwise pattern.
(b) RNN+ supports irregular time intervals, but the prediction trajectory still exhibits a stepwise pattern.
(c) NODE enable continuous modeling of irregular interval data, but the prediction trajectory deviates from the true values as the sequence length increases.
(d) NODE-RNN+ leverages the update ability of RNN+ to reduce the bias in fitting irregular interval data, but introduces the stepwise property of RNN.
(e) NODE-RGRU introduces an interpolation-driven continuous modeling method, further reducing the bias in fitting irregular interval data. however, interpolation can introduce noise, and the stepwise property of RNNs remains.
(f) The proposed model R-TNAG utilizes the update ability of the ARGRU for irregular interval data, and simultaneously incorporates dynamic time-aware parameters in NODE to smooth the stepwise behavior of the GRU model.}
\label{fig:all}
\end{figure}

As shown in \cref{fig:all}, temporal observations $t_1, t_2, \dots, t_n$ are used to predict the disease state and clinical scores at $t_{n+1}$, making robust temporal fitting under irregular intervals essential. To model AD progression from irregularly sampled longitudinal data, prior studies can be broadly categorized into five research directions: (a) Recurrent Neural Networks (RNN) \cite{rnn,lstm,gru,rnnlstmgru,rnn_lstm_gru2019}; (b) RNN irregular-interval extensions (RNN+) \cite{Liu2024,ALLN2024,TITD2025,DNA-T,MST_tr,TA-RNN2024ta}; (c) Neural Ordinary Differential Equations (NODE) \cite{ode,Bilos2021,2023LSS_NODE,2023node_eyes}; (d) NODE–RNN+  \cite{rubanova2019latent,ode2019gru,LaTiM}; and (e) Neural Ordinary Differential Equations and Riemannian manifold Gated Recurrent Unit (NODE-RGRU) \cite{odergru_ad}. However, methods in \cref{fig:all}(a)–(b) rely on discrete updates: RNN requires uniformly sampled data and thus depends on interpolation, which introduces errors and compromises prediction accuracy \cite{jing2025sm}, while RNN+ handles irregular intervals but still produces stepwise trajectories that fail to represent the continuous disease process.
To overcome this limitation, NODE-based approaches \cref{fig:all}(c) perform continuous-time modeling but are prone to inference drift due to error accumulation. NODE–RNN+ \cref{fig:all}(d) partially alleviates this issue by incorporating RNN+ updates; yet, their Euclidean feature space limits the expressiveness of temporal representations. NODE-RGRU in \cref{fig:all}(e) maps hand-crafted regional features into a Riemannian manifold, better capturing high-order relationships. However, it still depends on manually designed inputs that overlook fine-grained morphological changes.
To address these limitations, we propose the framework in \cref{fig:all}(f), which directly maps 3D sMRI features into a manifold space and employs an Attention-based Riemannian Gated Recurrent Unit (ARGRU) within a Time-aware Neural Ordinary Differential Equation (TNODE) model, enabling smooth, continuous trajectory learning under irregular sampling and accurate tracking of disease progression.

To address the challenge of AD prediction under irregularly sampled longitudinal 3D sMRI data, we develop an end-to-end spatio-temporal framework: a Riemannian manifold mapping, a Time-aware manifold Neural ordinary differential equation, and an Attention-based riemannian Gated recurrent unit (R-TNAG) that jointly learns spatial representations and models temporal dynamics. The main contributions of this study are as follows.

(1)End-to-end manifold feature extraction: We introduce a feature extraction pipeline that directly derives spatial features from 3D sMRI, employs a channel attention mechanism to suppress redundancy in high-dimensional representations, and maps the resulting features into a manifold space. This design enables the model to capture subtle morphological changes associated with disease progression.

(2)TNODE: We develop a TNODE module that constructs a time-aware NODE in manifold space, enabling continuous time modeling that alleviates the stepwise updates of ARGRU and produces smoother trajectories aligned with disease progression.

(3)ARGRU with interval-dependent scaling: We propose a modified Gated Recurrent Unit (GRU) update gate in manifold space, incorporating a time interval–dependent scaling factor. This enhancement improves the model’s ability to fit irregular-interval sequences and prevents deviation in prediction trajectories as observation intervals increase.

\section{Related Work}
\label{sec:related_work}
\subsection{Feature Extraction from Structural MRI}
The sMRI is widely used in clinical practice due to its non-invasive nature and relatively low cost \cite{alzheimer20192019}. A key challenge in AD progression modeling is to extract spatio-temporal deformation features from sMRI that accurately reflect disease development. Current methods fall into two main categories: single-time point analysis and longitudinal modeling.

Single-time point approaches focus on structural feature extraction from sMRI acquired at a single visit. For example, Yagis et al. \cite{3dVGG} employs a 3D Visual Geometry Group (VGG) variant to mitigate information loss from 2D convolutional filters, while UNEt TRansformers (UNETR++)\cite{unetr++} introduce a paired-attention mechanism that improves 3D medical image representation with a lower computational cost. These methods achieve strong performance in capturing structural patterns but largely ignore the temporal dependencies that are critical for modeling disease progression.

To incorporate temporal dynamics, longitudinal approaches leverage multi-visit sMRI, aggregating spatial features across time points into unified representations \cite{nolte2025cnn} or encoding multi-time point data into a latent space to reconstruct potential trajectories \cite{fu2025synthesizing}. While these methods enhance temporal modeling, the high dimensionality and complex spatial structure of longitudinal sMRI often lead deep networks to capture redundant or irrelevant representations, ultimately degrading predictive performance.

To address these challenges, we propose an end-to-end framework that directly extracts spatial features from 3D sMRI, applies channel attention to suppress redundancy and emphasize task-relevant information, and maps the refined representations into a Riemannian manifold space to more effectively model subtle morphological changes associated with AD progression.

\subsection{Temporal Continuity Modeling in Longitudinal Data}

Traditional discrete-time models struggle to capture the inherently continuous evolution of disease processes. To address this limitation, recent studies have proposed continuous-time dynamic modeling approaches based on temporal data, implemented either in Euclidean space or on Riemannian manifolds.

In Euclidean space, several works \cite{ode,rubanova2019latent,LaTiM} have extended NODE-based methods to achieve continuous trajectory modeling. NODE \cite{ode} parameterizes the derivative of hidden states with respect to time, enabling continuous trajectory modeling but relying heavily on the initial condition, which leads to prediction drift as observations accumulate. To reduce this drift, ODE-RNN \cite{rubanova2019latent} combines NODE integration between observations with hidden-state updates at observation points via RNN; yet it still lacks explicit temporal awareness. Building on this, Zeghlache et al. \cite{LaTiM} introduced time-aware NODE heads and a contrastive learning framework to better regularize latent trajectories. Despite these improvements, Euclidean approaches generally fail to capture feature dependencies in a geometrically meaningful way.

To overcome this, manifold-based methods have been explored. Jeong et al.\cite{odergru_old,odergru_ad} proposed NODE-RGRU, which embeds NODE in a Riemannian manifold, such as the space of symmetric positive-definite matrices, allowing for richer modeling of feature correlations. However, the recurrent updates in RGRU still introduce stepwise behavior at observation points, compromising the smoothness of predicted trajectories.

Thus, although continuous-time approaches improve the modeling of irregularly sampled data, learning smooth trajectories on manifolds remains challenging. We address this by introducing a TNODE framework that produces continuous and stable predictions of disease progression.

\subsection{Modeling Irregularly Sampled Time Series}

A major challenge in AD progression modeling is capturing temporal dynamics from irregularly sampled longitudinal data. Traditional RNN-based methods for disease prediction \cite{jung2021deep}, such as GRU \cite{park2024predicting} and Long Short-Term Memory (LSTM) \cite{rnn_lstm_gru2019}, typically address this issue by imputing missing data to regularize input sequences. Although this ensures uniform temporal spacing for model training, it can introduce imputation errors and increase computational overhead. To overcome these limitations, recent studies explore methods that directly model irregular intervals, which can be broadly classified into fixed-parameter and dynamic-parameter approaches.

Fixed-parameter designs, such as Gated Recurrent Units with Decays (GRU-D) \cite{GRU-Dche2018recurrent} and a temporal information enhancing LSTM (T-LSTM) \cite{T-LSTM2019t}, extend RNN structures with pre-defined temporal adjustments. for example, they introduce decay factors or separately model memory decay and updates from new inputs. Although these approaches improve temporal adaptability, their parameters are proportionally tied to time intervals and may fail to capture the nonlinear and heterogeneous progression of AD.

Dynamic-parameter methods adjust update rules at each observation point to better capture irregular temporal dynamics. The GRU-ODE-Bayes\cite{ode2019gru} combines GRU-ODE and GRU-Bayes, extends GRU for irregular intervals but can suffer from prediction drift, while the NODE-RGRU series \cite{odergru_old,odergru_ad} redesigns GRU in manifold space and, in \cite{odergru_ad}, introduces a trajectory prediction module for AD. 
Nevertheless, NODE-RGRU methods often rely on interpolation, incurring extra costs and errors, which underscores the need for more flexible mechanisms to capture heterogeneous temporal variations.

Despite these advances, modeling heterogeneous temporal variations under irregular intervals remains challenging.
To address this, our method R-TNAG employs a manifold-based ARGRU with interval-dependent scaling to adaptively capture irregular dynamics and stabilize trajectories.

\section{Method} 
\label{sec:method}

To address the challenge of AD prediction under irregular temporal intervals, we propose R-TNAG, an end-to-end framework that integrates feature manifold mapping, continuous-time dynamics, and irregularity-aware mechanisms. By jointly modeling longitudinal sMRI and cognitive scores, R-TNAG enables accurate forecasting of AD progression.

\begin{figure*}[t]	
\centerline{\includegraphics[width=\linewidth]{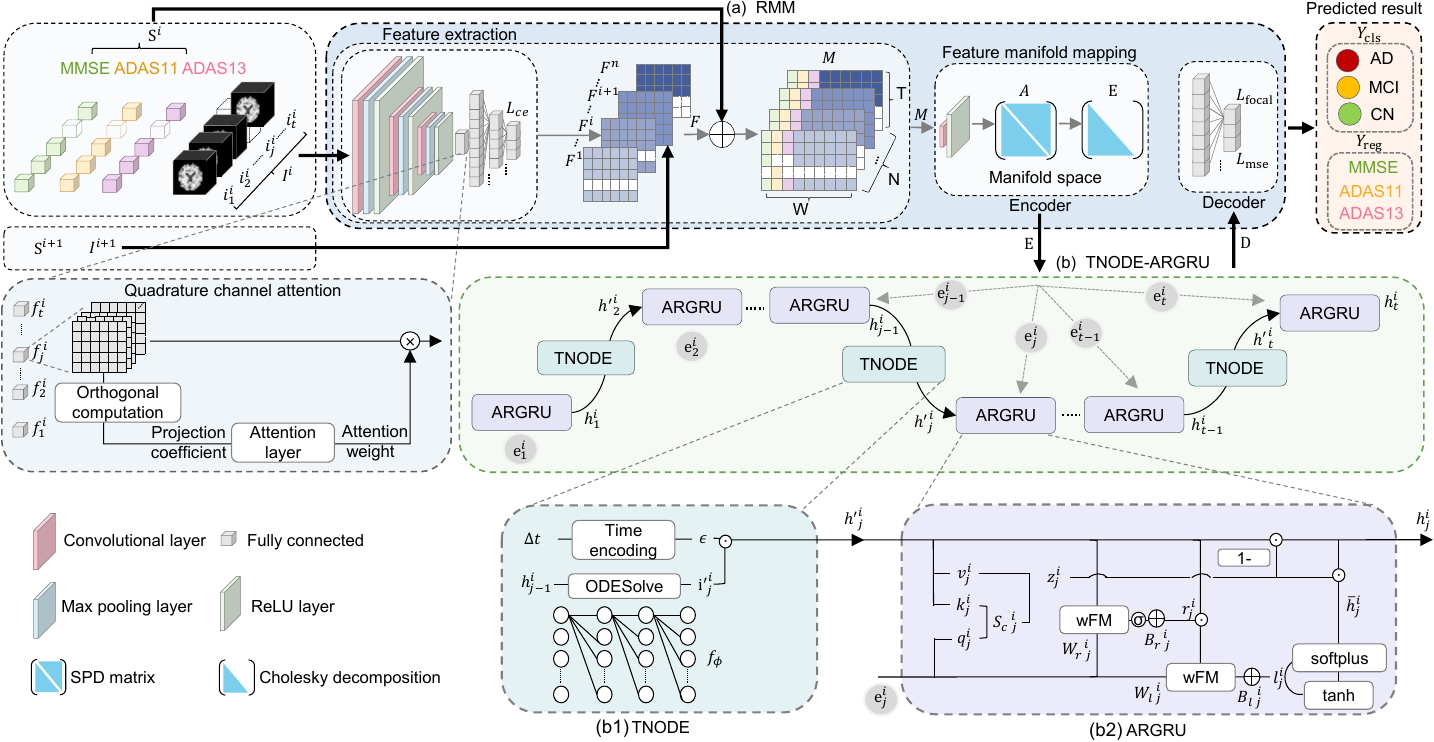}}
\caption{
The overall architecture of the R-TNAG model is shown, where longitudinal clinical scores and imaging sequences serve as inputs, and the outputs are predictions of diagnostic categories and cognitive scores. (a) The RMM module is responsible for high-dimensional feature extraction and manifold embedding. (b) The TNODE-ARGRU module performs disease progression modeling by alternating between TNODE and ARGRU submodules; (b1) the TNODE submodule is a time-aware NODE that continuously models state evolution over time; (b2) the ARGRU submodule is an attention-augmented GRU that refines trajectory updates at irregular observation intervals.
}
\label{fig:model}
\end{figure*}

As illustrated in \cref{fig:model}, the proposed R-TNAG framework comprises two main components: a Riemannian Manifold Mapping (RMM) module and a Time-aware Neural ODE with Attention-based Riemannian GRU (TNODE-ARGRU) module.
In \cref{fig:model}(a), the RMM module performs feature extraction and feature manifold mapping. Extracted sMRI features are concatenated with cognitive assessments, including the Mini-Mental State Examination (MMSE), Alzheimer’s Disease Assessment Scale–Cognitive Subscale 11 (ADAS11), and Alzheimer’s Disease Assessment Scale–Cognitive Subscale 13 (ADAS13), and projected onto a manifold of Symmetric Positive Definite (SPD) matrices, thereby preserving the intrinsic geometric structure of disease progression.
Subsequently, as shown in \cref{fig:model}(b), the TNODE-ARGRU module alternates between TNODE and ARGRU operations to model temporal evolution. The TNODE submodule implements a time-aware neural ordinary differential equation in the manifold space to continuously model the evolution of latent states across varying time intervals, while the ARGRU submodule is a gated unit with an attention-based update mechanism in manifold space, enhancing the model’s ability to handle irregularly spaced temporal sequences.

\subsection{Definitions and Notation}
\subsubsection{Notation}
In this subsection, $N$ and $T$ denote the number of subjects and time points, respectively. $Z$ represents the dimensionality of sMRI data, and $K$ represents the dimensionality of cognitive measures. $O$ denotes the dimensionality of extracted features, $W$ denotes the dimensionality of multimodal features, and $Q$ denotes the dimensionality of manifold  features. $C$ is the number of diagnostic categories, and $R$ is the dimensionality of regression targets. The indices $i \in {1,\dots,N}$ and $j \in {1,\dots,T}$ indicate the $i$-th subject at time point $j$.

\subsubsection{Inputs}
The longitudinal 3D sMRI sequence is defined as 
$I=\{I^1,I^2,...I^i...,I^{n-1},I^n\}\in \mathbb{R}^{N \times T \times Z}$,where 
$I^i=\{i^{i}_{1},i^{i}_{2},\dots,i^{i}_{j},\dots,i^{i}_{t-1},i^{i}_{t}\}\in \mathbb{R}^{T \times Z}$, and $i^{i}_{j} \in \mathbb{R}^{Z}$. 
The cognitive score sequence is defined as $S=\{S^1,S^2,...S^i...,S^{n-1},S^n\}\in \mathbb{R}^{N \times T \times K}$,where 
$S^i=\{s^{i}_{1},s^{i}_{2},\dots,s^{i}_{j},\dots,s^{i}_{t-1},s^{i}_{t}\}\in \mathbb{R}^{T \times K}$, and $s^{i}_{j} \in \mathbb{R}^{K}$. 

\subsubsection{Feature Extraction}
The output of the feature extraction network is defined as $F=\{F^1,F^2,...F^i...,F^{n-1},F^n\}\in \mathbb{R}^{N \times T \times O}$,where 
$F^i=\{f^{i}_{1},f^{i}_{2},\dots,f^{i}_{j},\dots,f^{i}_{t-1},f^{i}_{t}\}\in \mathbb{R}^{T \times O}$, and $f^{i}_{j} \in \mathbb{R}^{O}$. The multimodal feature set is defined as $M=\{m^{1}_{1},m^{1}_{2},\dots,m^{i}_{j},\dots,m^{n}_{t-1},m^{n}_{t}\} \in \mathbb{R}^{N\times T \times W}$, where $m^{i}_{j} \in \mathbb{R}^{W}$.  

\subsubsection{Encoder and Decoder}
The encoder output is $E=\{e^{1}_{1},e^{1}_{2},\dots,e^{i}_{j},\dots,e^{n}_{t-1},e^{n}_{t}\} \in \mathbb{R}^{N\times T \times Q}$, where $e^{i}_{j} \in \mathbb{R}^{Q}$, which also serves as the decoder input $D=\{d^{1}_{1},d^{1}_{2},\dots,d^{i}_{j},\dots,d^{n}_{t-1},d^{n}_{t}\} \in \mathbb{R}^{N\times T \times Q}$, where $d^{i}_{j} \in \mathbb{R}^{Q}$.

\subsubsection{Hidden States}
The hidden states of TNODE are $H^\prime=\{{h^\prime}^{1}_{1},{h^\prime}^{1}_{2},\dots,{h^\prime}^{i}_{j},\dots{{h^\prime}^{n}_{t-1}},{{h^\prime}^{n}_{t}}\} \in \mathbb{R}^{N\times T \times Q}$, where ${h^\prime}^{i}_{j} \in \mathbb{R}^{Q}$. While the hidden states of the ARGRU are $H=\{h^{1}_{1},h^{1}_{2},\dots,h^{i}_{j},\dots,h^{n}_{t-1},h^{n}_{t}\} \in \mathbb{R}^{N\times T \times Q}$, where $h^{i}_{j} \in \mathbb{R}^{Q}$.  

\subsubsection{Outputs}
The output of the diagnostic classification task is  
$
Y_{cls} = \{ y^1_{1, cls},y^1_{2, cls},\dots,y^i_{j, cls},\dots,y^n_{t, cls} \} \in \mathbb{R}^{N \times T \times C}
$, where $y^{n}_{t,cls} \in \mathbb{R}^{C}$.   
The output of cognitive score regression is $
Y_{reg} = \{ y^1_{1, reg},y^1_{2, reg}\dots,y^i_{j, reg},\dots,y^n_{t, reg} \} \in \mathbb{R}^{N \times T \times R}$, where $y^{n}_{t,reg} \in \mathbb{R}^{R}$. The overall model output is $Y=\{Y_{cls},Y_{reg}\}$.

\subsection{RMM}\label{RMM}

To capture disease-relevant geometric variations from 3D sMRI, the RMM module employs a channel-attention feature extractor and maps the resulting representations into a manifold space.
As illustrated in \cref{fig:model}(a), the RMM module first processes the 3D sMRI $I$ through a feature extraction block to obtain feature maps $F$. These are concatenated with cognitive score data $S$ to form multimodal features $M$, which are then projected onto the manifold space to yield the representation $A$. From $A$, the lower-triangular matrix $E$ is derived and subsequently fed into the TNODE-ARGRU module in \cref{fig:model}(b) to produce the final prediction $D$.

\subsubsection{Feature extraction}
To obtain effective disease-related features from whole brain images, the RMM module transforms the complex and high-dimensional 3D sMRI data into low-dimensional features through a feature extraction block, laying the foundation for subsequent manifold space modeling.
The feature extraction block in \cref{fig:model}(a) consists of three convolution–pooling units, each containing a $3 \times 3 \times 3$ convolution, a $2 \times 2 \times 2$ max-pooling operation, and a ReLU activation. After the final activation, the high-dimensional image is converted into a low-resolution, high-channel feature map. To mitigate channel redundancy and highlight disease-relevant patterns, a Gram–Schmidt-based channel attention mechanism is applied prior to the fully connected layers. This mechanism computes orthogonal projection coefficients across channels, passes them through a two-layer attention network to generate adaptive channel weights, and reweights the original feature maps to enhance salient features while suppressing irrelevant information. The reweighted features are then flattened and passed through two fully connected layers: the first produces imaging features $F$ for downstream tasks, and the second is optimized with the cross-entropy loss $L_{ce}$ in \eqref{eq:ce}, guided by disease-state classification.

\begin{equation}
\label{eq:ce}
L_{ce} = - \sum_{c=1}^{C} \| P_c(F) \odot \log Q_c(F) \|_1,
\end{equation}
Here, $P_c(F)$ denotes the ground-truth probability distribution for class $c$, and $Q_c(F)$ denotes the model-predicted distribution for class $c$. The operator $\odot$ represents element-wise multiplication, and $|\cdot|_1$ indicates summation over all elements. The number of classes is $C = 3$, corresponding to Cognitively Normal (CN), Mild Cognitive Impairment (MCI), and AD.

\subsubsection{Feature manifold mapping}

To preserve intrinsic geometric relationships and more effectively capture disease trajectories, the feature manifold mapping (FMM) projects multimodal representations onto a Riemannian manifold, as illustrated in \cref{fig:model}(a) within the RMM module
Specifically, imaging features $F$ are concatenated with cognitive scores $S$ to form multimodal representations $M$, which are then processed by a 1D convolution to increase channel dimensionality. The covariance matrix of $M$ is subsequently computed, naturally yielding an SPD matrix that resides on a convex Riemannian manifold, denoted as $A$. To enhance computational efficiency, Cholesky decomposition factorizes the SPD matrix into a lower triangular matrix and its transpose, producing the manifold representation $E$, which is passed to the TNODE-ARGRU module to generate $D$.
Finally, a task-specific decoder comprising dense layers maps $D$ to outputs for two prediction tasks. The disease state prediction task, optimized with the focal cross-entropy loss $L_{focal}$ in \eqref{eq:focal}, drives the model to capture disease-relevant features from MRI data. In parallel, the cognitive score regression task, trained with the mean squared error loss $L_{mse}$ in \eqref{eq:lmse}, provides complementary supervision that stabilizes feature learning and improves robustness.

\begin{equation}
\label{eq:focal}
L_{focal} = - \sum_{c=1}^{C} \alpha_c \, \| P_c(Y_{cls}) \odot (1 - Q_c({Y_{cls}))^\gamma \odot \log Q_c(Y_{cls}}) \|_1
\end{equation}
Here, $\alpha_c$ and $\gamma$ are hyperparameters controlling the class weight and focal modulation, respectively. $P_c(Y_{cls})$ denotes the ground-truth probability for class $c$, and $Q_c(Y_{cls})$ denotes the model-predicted probability for class c. The symbol $\odot$ represents element-wise multiplication, and $|\cdot|_1$ indicates summation over all elements. The number of classes is $C=3$, corresponding to CN, MCI, and AD.

\begin{equation}
\label{eq:lmse}
L_{mse} = \frac{1}{K} \| E \odot B - Y_{reg} \odot B \|^2_F
\end{equation}
Here, $E$ and $Y_{reg}$ denote the ground-truth cognitive scores and the model’s regression predictions, respectively. $B$ is a mask matrix, $\odot$ represents element-wise multiplication, $|\cdot|_F$ denotes the Frobenius norm, and $K$ is the total number of valid observation points where the mask equals 1.

\subsection{TNODE-ARGRU}\label{DPPM}      

To capture geometric relationships of brain morphology from 3D sMRI and predict disease progression under irregular intervals, the TNODE-ARGRU architecture is shown in \cref{fig:model}(b). Operating entirely in manifold space, it alternates between TNODE and ARGRU components to model temporal dynamics.
For an input sequence $E$, at each observation time $j$, TNODE evolves the hidden state from the previous step, denoted as ${{h^\prime}^{i}_{j}}$. Given the current observation $e^{i}_{j}$ and ${{h^\prime}^{i}_{j}}$, the ARGRU updates the hidden state $h^{i}_{j}$, which is then fed into TNODE for the next time point.

The algorithm is summarized in \cref{algorithm1}. The inputs are $h_{j-1}^i$, $f_\Phi$, $\Phi$, and $\epsilon$, and the output is ${{h}^{i}_{j}}$, where $h^i_{j-1}$ is the previous hidden state, $f_\Phi$ is the neural network, $\Phi$ are its parameters, $\epsilon$ is the time coefficient, and ${{h}^{i}_{j}}$ is the updated hidden state. TNODE and ARGRU are executed iteratively in an alternating manner.

\begin{algorithm}
\caption{TNODE-ARGRU}
\label{algorithm1}
\textbf{Input:}{$h_{j-1}^i$, $f_\Phi$, $\Phi$ $\epsilon$, $e^{i}_{j}$} 
\\
\textbf{Output:}{${{h}^{i}_{j}}$}

\begin{algorithmic}[1]  
    \State Initial hidden state: $h_{0}^i$ 
    \For{$i \in \{ 1, \ldots, n\}$}
        \For{$j \in \{ 1, \ldots, t\}$}
         \State //Calculate TNODE
        \State ${{h^\prime}^{i}_{j}} \leftarrow$ Numerical integration$(h_{j-1}^i,f_\Phi,\Phi,\epsilon)$ 
       \State //Calculate ARGRU
        \State $z^{i}_{j}\leftarrow$ Calculate attention update gate $({S_c},v^{i}_{j})$
        \State$r^{i}_{j}\leftarrow$ Calculate the reset gate$({e^{i}_{j}, h^\prime}^{i}_{j})$
        \State$\bar{h^{i}_{j}}\leftarrow$ Calculate candidate hidden state$(l^{i}_{j})$
        \State${h^{i}_{j}}\leftarrow$ Calculate current hidden state$(
        {z^{i}_{j},{h^\prime}^{i}_{j}},\bar{h^{i}_{j}})$
        \EndFor
    \EndFor

    \State \Return Updated hidden state ${{h}^{i}_{j}}$
\end{algorithmic}
\end{algorithm}

\subsubsection{TNODE}\label{T-NODEs}

Modeling continuous disease progression from discrete AD observations requires capturing latent dynamics between discrete time points. To this end, the TNODE submodule leverages a time-aware NODE to compute the continuous evolution of hidden states.

As shown in \cref{algorithm1}, TNODE takes four parameters as input: $h_{j-1}^i$, $f_\Phi$, $\Phi$, and $\epsilon$. 
The initial hidden state $h_{0}^i$ of TNODE is formed by an identity matrix. Finally, the state ${{h^\prime}^{i}_{j}}$ at time $j$ is obtained through a solver.  

The TNODE module takes the hidden state $h_{j-1}^i$, produced by the ARGRU at the previous time point, as input and introduces a learnable time coefficient $\epsilon$ to dynamically modulate the effect of varying time intervals on the rate of continuous modeling. Within each interval $[j-1, j]$, the neural network $f_\Phi$ within the TNODE solver computes the transition from $h_{j-1}^i$ to $h_j^i$ via numerical integration.
During integration, patient age is passed through a time-encoding layer to produce the learnable coefficient $\epsilon$, which scales the hidden state within the differential solver. This enables temporal information to be fused into the hidden dynamics, allowing the model to adaptively learn disease progression rates. The coefficient $\epsilon$ effectively governs the rate of change in the neural ordinary differential equation, making it dependent not only on the previous state but also on the interval length. For the initial time, the initialization of the hidden state $h_{0}^i$ is computed in \eqref{eq:h0},
and the hidden state ${h^\prime}^{i}_{j}$, as defined in \eqref{eq:h'}, is taken as the output of the TNODE module.

\begin{equation}
\label{eq:h0}
h_{0}^i = h^i(j=0)
\end{equation}
where $h_{0}^i$ is the initial identity matrix.

\begin{equation}
\label{eq:h'}
{{h^\prime}^{i}_{j}}=h_{j-1}^i+\int_{j-1}^{j} \epsilon f_\Phi(j, h^i(j), \Phi) \ dj
\end{equation}

\subsubsection{ARGRU}\label{ARGRU}

To handle the long time spans and irregular intervals characteristic of AD diagnostic data, we adopt the ARGRU module. Within the gated recurrent units, the update gate plays a critical role in balancing historical information with new input, ensuring that the hidden state reflects both long-term dependencies and recent observations. 

The overall procedure of ARGRU is summarized in Algorithm~\ref{algorithm1}.
The input to this module consists of the hidden state ${{h^\prime}^{i}_{j}}$ propagated from the TNODE module and the current multimodal feature $e^{i}_{j}$. An attention mechanism is employed to model temporal dependencies, while the reset gates $r^{i}_{j}$ and attention update gates $z^{i}_{j}$ jointly regulate the hidden state. This design enables adaptive integration of past information with the current observation, ensuring that the updated hidden state $h^{i}_{j}$ effectively reflects both historical context and new input.

The attention update gate $z^{i}_{j}$ is defined in \eqref{eq:z}-\eqref{eq:v} and computed by weighting the value vector $v^{i}_{j}$ with the attention score $S_c$, which quantifies the relevance of input features and enables the model to selectively extract task-relevant information. The query vector $q^{i}_{j}$  is obtained by mapping the hidden state ${{h^\prime}^{i}_{j}}$ with $W_{q}$, while the key $k^{i}_{j}$ and value $v^{i}_{j}$ vectors are derived from the multimodal input $e^{i}_{j}$ via $W_{k}$ and $W_{v}$, respectively. This attention-guided gating mechanism allows the query to locate the most relevant historical context, balance past and current information, and improve the capture of temporal dependencies in disease progression.

\begin{equation}  
\label{eq:z}
 z^{i}_{j} = {S_c}v^{i}_{j}
\end{equation}
\begin{equation} 
\label{eq:sc}
 S_c = \text{softmax}\left( \frac{q^{i}_{j} {k^{i}_{j}}^T}{\sqrt{d_k}} \right)
\end{equation} 
\begin{equation} 
\label{eq:q}
q^{i}_{j} = \exp(W_{q} \log({{h^\prime}^{i}_{j}}))
\end{equation}
\begin{equation}
\label{eq:k}
k^{i}_{j} = \exp(W_{k} \log(e^{i}_{j}))
\end{equation}
\begin{equation}
\label{eq:v}
v^{i}_{j} = \exp(W_{v}\log(e^{i}_{j}))
\end{equation}  
where, softmax$(\cdot)$ is a softmax function. $d_k$ is the dimensionality of the key. $W_q$, $W_k$, and $W_v$ are learnable weight matrices. 

The reset gate $r^{i}_{j}$ is computed according to \eqref{eq:ri}, which controls how much past information is preserved. Using $r^{i}_{j}$, the intermediate state $l^{i}_{j}$ is obtained via \eqref{eq:li}, and the candidate hidden state $\bar{h^{i}_{j}}$ is then computed as in \eqref{eq:hbar}. Finally, the current hidden state $h^{i}_{j}$ is updated by combining $\bar{h^{i}_{j}}$ with the previous hidden state, as formulated in \eqref{eq:h}.

\begin{equation}
\label{eq:ri}
r^{i}_{j}=\sigma({wFM}(\{e^{i}_{j}, {{h^\prime}^{i}_{j}}\}, W_r)\oplus B_r \end{equation}
\begin{equation}
\label{eq:li}
l^{i}_{j}={wFM}(\{e^{i}_{j}, r^{i}_{j} \oplus {{h^\prime}^{i}_{j}}\}, W_l) \oplus B_l
\end{equation}
\begin{equation}
\label{eq:hbar}
\bar{h^{i}_{j}} = \tanh(L({l^{i}_{j}})) + \text{softplus}(D({l^{i}_{j}}))
\end{equation}
\begin{equation}
\label{eq:h}
h^{i}_{j} = (1 - z^{i}_{j}) \odot {{h^\prime}^{i}_{j}} + z^{i}_{j} \odot \bar{h^{i}_{j}}
\end{equation}
Where $wFM$ denotes the Fréchet Mean \cite{Frchet1948LesA}, and $\sigma(\cdot)$ represents the logistic sigmoid function. $W_r$ and $W_l$ are learnable weight matrices, while $B_r$ and $B_l$ are their corresponding learnable biases. The operation $\oplus$ denotes a smooth commutative group operation on the manifold \cite{odergru_old}. The functions $\tanh(\cdot)$ and softplus $(\cdot)$ denote the hyperbolic tangent and softplus functions, respectively. $L(\cdot)$ and $D(\cdot)$ represent the lower-triangular and diagonal matrices obtained from Cholesky decomposition, respectively. $h^{i}_{j}$ is the hidden state at time $j$, and ${{h}^{i}_{j}}$ is the final output of ARGRU.

\section{Experiments} 
\label{sec:xperiments}
\subsection{Dataset}
\subsubsection{Data preprocessing}

The dataset was derived from the Alzheimer's Disease Neuroimaging Initiative (ADNI). Participants with only baseline scans or fewer than three follow-ups were excluded, while those with 1–5 years of longitudinal visits were retained.

Table~\ref{tab:distribution} shows the diagnostic distribution across follow-up periods. Most records are concentrated at baseline (BL) and within the first two years (M24), with a marked decline thereafter (M36-M60). Missing rates rise accordingly, from 16.9\% at 1 year to 30–39\% between 2 and 5 years, reaching 39.2\% at 5 years, which serves as the main setting for our experiments.
For imaging preprocessing, raw sMRI scans were skull-stripped with HD-BET \cite{HBBET}, orientation-corrected and intensity-normalized using FSL \cite{FSL}, registered to MNI152, and resampled to 180×180×180. Cognitive scores (MMSE, ADAS11, ADAS13) were normalized to [0,1].

\begin{table}
\centering
\caption{Distribution of Diagnostic Categories Over Different Follow-Up Periods in ADNI
}
\label{tab:distribution}
    \resizebox{0.95\columnwidth}{!}{
\begin{tabular}{llccccccccc}
\toprule
\multirow{2}{*}{Dataset} & \multirow{2}{*}{Label} & \multicolumn{9}{c}{Follow-up Periods} \\
\cmidrule(lr){3-11}
                         &                        & BL  & M06 & M12 & M18 & M24  & M36  & M48 & M54 & M60  \\
\midrule
\multirow{3}{*}{ADNI}     & CN   & 402  & 377 & 401 & 9 & 348  & 152  & 167 & 0 & 52 \\
                         & MCI  & 760  & 699 & 639 & 196 & 432  & 177  & 159 & 1 & 35 \\
                         & AD   & 248  & 281 & 332 & 82 & 285  & 102 &  95 & 0 & 8 \\
\bottomrule
\end{tabular}
}
\end{table}

\subsubsection{Experimental Setup}
All experiments in this subsection were conducted on a workstation equipped with four NVIDIA TITAN Xp GPUs (12 GB memory each), using PyTorch with CUDA support for model training and inference. A five-fold cross-validation strategy was adopted, with an 80\%–20\% split between training and testing sets. Models were trained for 300 epochs using the Adam optimizer with a learning rate of 0.001.
For the disease status prediction task, evaluation metrics included the mean area under the receiver operating characteristic curve (mAUC), Precision, Recall, and F1-score (F1). For the cognitive score regression task, the mean absolute percentage error (MAPE) and the coefficient of determination ($R^2$) were used as evaluation metrics.

\subsection{Ablation Study}

\subsubsection{Ablation Study of Feature Extraction in RMM}

To assess the effectiveness of the RMM module, an ablation study was performed by replacing it with alternative feature extraction models and comparing the resulting disease state prediction performance over 5 years of longitudinal data. Specifically, the RMM module was substituted with several representative 3D medical image feature extractors, including VGG3D for AD classification \cite{3dVGG}, UNETR++ which demonstrates strong performance in medical image segmentation \cite{unetr++}, and the feature extraction components of longitudinal imaging networks such as HCCNet \cite{nolte2025cnn} and InBrainSyn \cite{fu2025synthesizing}. All replacement models were configured to produce output features of the same dimensionality and were integrated with the TNODE-ARGRU module for a fair comparison.

\begin{table*}[!t]
    \centering
    \caption{ABLATION EXPERIMENTS OF DIFFERENT FEATURE EXTRACTION MODULES}
    \label{table:FE_ablation}
    \resizebox*{0.9\textwidth}{!}{
\begin{tabular}{l|c|cccc}
\toprule
Feature extraction      &Classifier& mAUC$\uparrow$ & Precision$\uparrow$ & Recall$\uparrow$&F1$\uparrow$            \\  \midrule

VGG3D\cite{3dVGG} &\multirow{5}{*}{TNODE-ARGRU}& 0.8190 ± 0.0410  & 0.5452 ± 0.1018 & 0.5616 ± 0.1919 & 0.5145 ± 0.1229 \\
UNETR++\cite{unetr++}    && 0.7621 ± 0.0504 & 0.4327 ± 0.0441 & 0.5864 ± 0.1916 & 0.4422 ± 0.0387 \\
HCCNet\cite{nolte2025cnn} &&0.9427 ± 0.0078&0.8409 ± 0.0372& 0.8144 ± 0.0213 &0.8212 ± 0.0220  \\
InBrainSyn\cite{fu2025synthesizing}    &&0.8862 ± 0.0271  &0.7525 ± 0.0620  &0.6729 ± 0.0688  &0.6894 ± 0.0508  \\
RMM(Ours)    && \textbf{0.9582 ± 0.0224} & \textbf{0.8613 ± 0.0385} & \textbf{0.8411 ± 0.0450}  & \textbf{0.8479 ± 0.0441}\\
\bottomrule
\end{tabular}
}
\end{table*}

As shown in \cref{table:FE_ablation}, 3D convolutional models operating on single time points, namely VGG3D and UNETR++, demonstrated limited performance in temporal prediction tasks, with Precision, Recall, and F1 scores ranging from 0.4327 to 0.5864. In contrast, temporal modeling networks, such as HCCNet and InBrainSyn, achieved substantial improvements, with HCCNet reaching Precision, Recall, and F1 scores between 0.8144 and 0.8409. Notably, the proposed RMM consistently outperformed all baseline methods across all evaluation metrics, attaining the highest mAUC of 0.9582, along with superior Precision, Recall, and F1 scores. These results highlight RMM's strong capability in capturing and leveraging temporal imaging features for longitudinal prediction.

\subsubsection{Ablation Study of Spatial Modeling in TNODE-ARGRU}

To evaluate whether manifold-based modeling outperforms its Euclidean counterpart, we implemented TNODE-ARGRU modules with identical architectures, replacing all manifold-space operations with standard Euclidean computations. Instead of sMRI features extracted by the RMM module, we directly used manually curated one-dimensional MRI biomarkers and cognitive scores from 5-year patient data sequences as inputs to TNODE-ARGRU for prediction. The results are summarized in \cref{table:space_ablation}.

\begin{table}[H]
    \centering
    \caption{Ablation experiments with different spatial models}
    \label{table:space_ablation}
    \resizebox{\columnwidth}{!}{
    \begin{tabular}{lccccc}
    \toprule
        Space & mAUC$\uparrow$ & Precision$\uparrow$ & Recall$\uparrow$&F1$\uparrow$ \\ \midrule
        Euclidean & 0.7037 ± 0.1414 & 0.312 ± 0.0427 & 0.2264 ± 0.0681 & 0.2476 ± 0.0127 \\ 
        Manifold & \textbf{0.8972 ± 0.0135} & \textbf{0.7654 ± 0.0322} & \textbf{0.7227 ± 0.0626} & \textbf{0.7303 ± 0.0424} \\ \bottomrule
    \end{tabular}
    }
\end{table}

It can be observed that modeling in Manifold-space substantially outperformed Euclidean-space modeling across all evaluation metrics, including mAUC, Precision, Recall, and F1. Specifically, mAUC increased from 0.7037 to 0.8972, and F1 improved from 0.2476 to 0.7303, demonstrating that Manifold-space modeling markedly enhances the discriminative capability of temporal data. These findings confirm the advantage of manifold space in preserving intrinsic structural constraints and capturing complex dynamic features.


\subsubsection{Ablation Study of R-TNAG Modules}

To assess the contribution of each module, ablation experiments were performed on 5-year longitudinal data under five settings: (a) replacing RMM feature extraction with hand-crafted MRI biomarkers; (b) substituting TNODE with a conventional NODE in manifold space; (c) removing the ARGRU module; (d) replacing ARGRU with RGRU \cite{odergru_ad}; and (e) using the complete model. The results are summarized in \cref{table:ablation}.

\begin{table*}[!t]
\centering
\caption{Ablation Study of Model Components}
\label{table:ablation}
\resizebox*{0.9\textwidth}{!}{
    \begin{tabular}{lccccccc}
    \hline
        Case &  RMM & TNODE & ARGRU & mAUC$\uparrow$ & Precision$\uparrow$ & Recall$\uparrow$ & F1$\uparrow$ \\ \midrule
        (a) & × & $\checkmark$ &  $\checkmark$ & 0.8972 ± 0.0135 & 0.7654 ± 0.0322 & 0.7227 ± 0.0626 & 0.7303 ± 0.0424 \\ 
        (b) &  $\checkmark$ & × &  $\checkmark$ & 0.9514 ± 0.0351 & 0.8366 ± 0.0457 & 0.8333 ± 0.0791  & 0.8206 ± 0.0676 \\ 
        (c) &  $\checkmark$ &  $\checkmark$ & × & 0.9275 ± 0.0362 & 0.7290 ± 0.0619 & 0.6991 ± 0.0842 & 0.6348 ± 0.1231 \\ 
        (d) &  $\checkmark$ &  $\checkmark$ &  × & 0.9440 ± 0.0296 & 0.7699 ± 0.0700 & 0.8250 ± 0.0146 & 0.7583 ± 0.0718 \\ \midrule
        (e) &  $\checkmark$ &  $\checkmark$ &  $\checkmark$ & \textbf{0.9582 ± 0.0224} & \textbf{0.8613 ± 0.0385} & \textbf{0.8411 ± 0.0450} & \textbf{0.8479 ± 0.0441} \\ \bottomrule
\end{tabular}
}
\vspace{-0.1in}
\end{table*}

Compared with the complete model in Case (e), removing RMM in Case (a) reduced the mAUC from 0.9582 to 0.8972 and caused notable drops in Precision, Recall, and F1, highlighting the importance of high-dimensional manifold representations. Replacing TNODE with a conventional NODE in Case (b) slightly decreased all metrics, showing that continuous-time modeling alone cannot handle irregular intervals effectively. Removing ARGRU in Case (c) led to the largest decline, with F1 dropping from 0.8479 to 0.6348, emphasizing ARGRU role in integrating temporal features. Substituting ARGRU with RGRU \cite{odergru_ad} in Case (d) improved performance over complete removal but still reduced Precision to 0.7699 and F1 to 0.7583, confirming the need for explicit interval-aware modeling. Overall, Case (e) achieved the highest performance across all metrics.

These ablation results validate the necessity of our three innovations: RMM for capturing subtle 3D morphological changes, TNODE for continuous, interval-aware temporal modeling, and ARGRU with interval-dependent scaling for robust prediction on irregular sequences.

\subsection{Longitudinal Evaluation of Disease Status Prediction}

\begin{figure*}[t]
    \centering
    \subfloat{\includegraphics[width=0.24\linewidth]{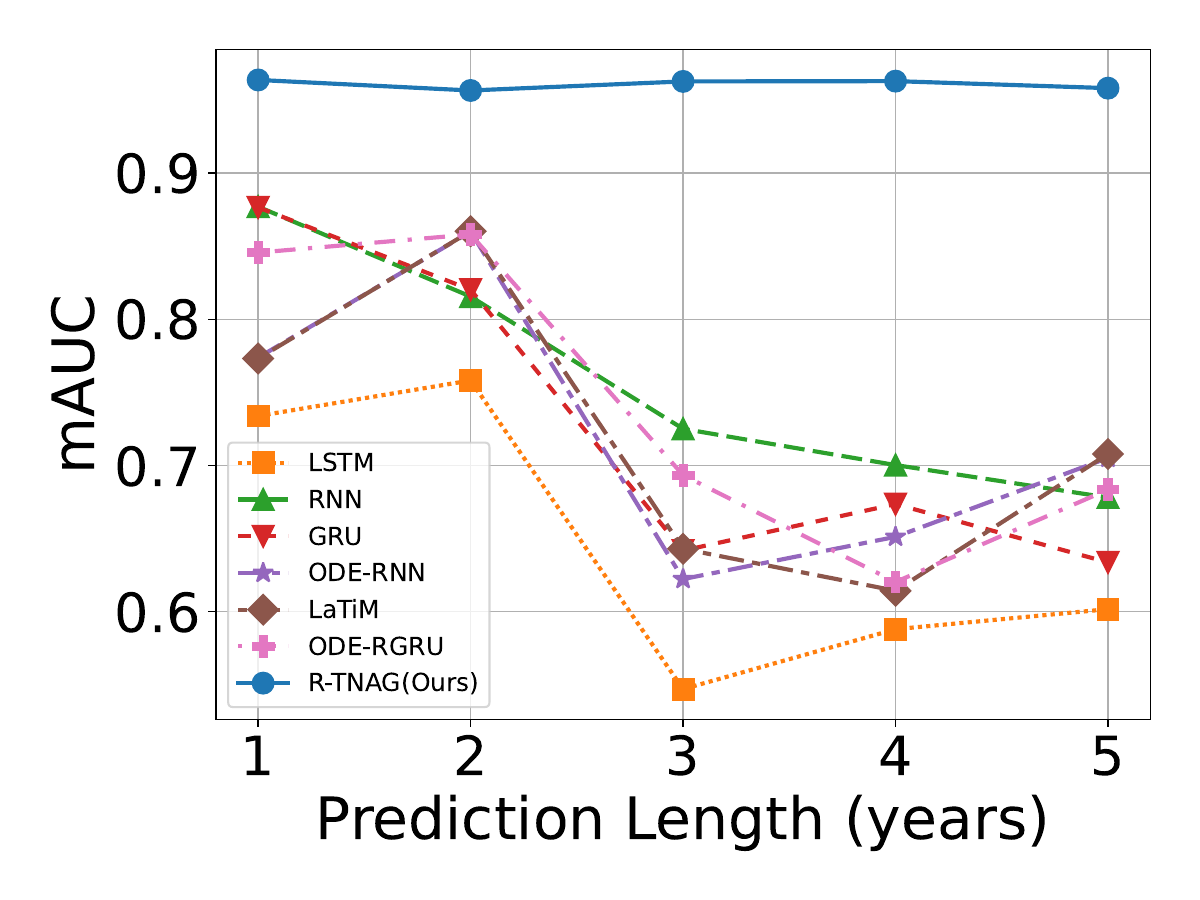}}
    \hfill
    \subfloat{\includegraphics[width=0.24\linewidth]{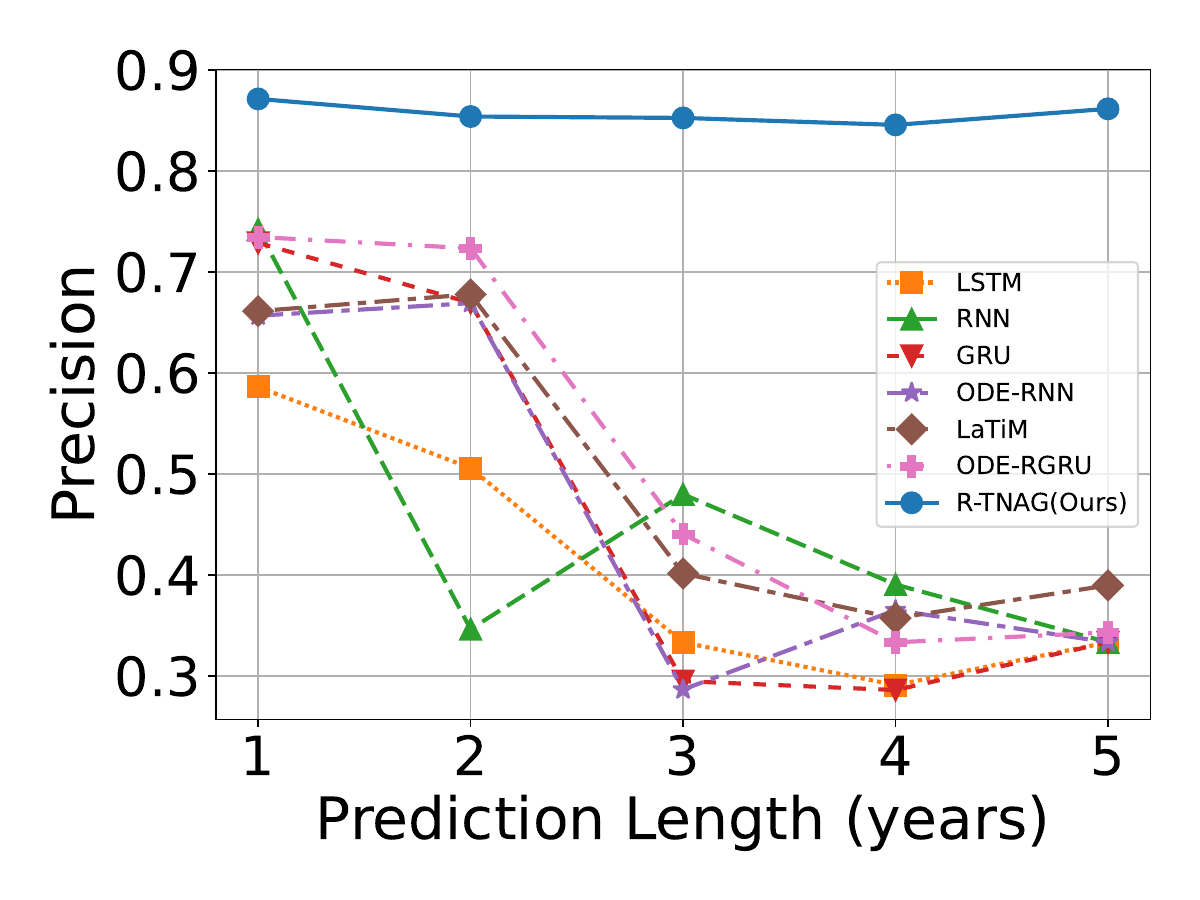}}
    \hfill
    \subfloat{\includegraphics[width=0.24\linewidth]{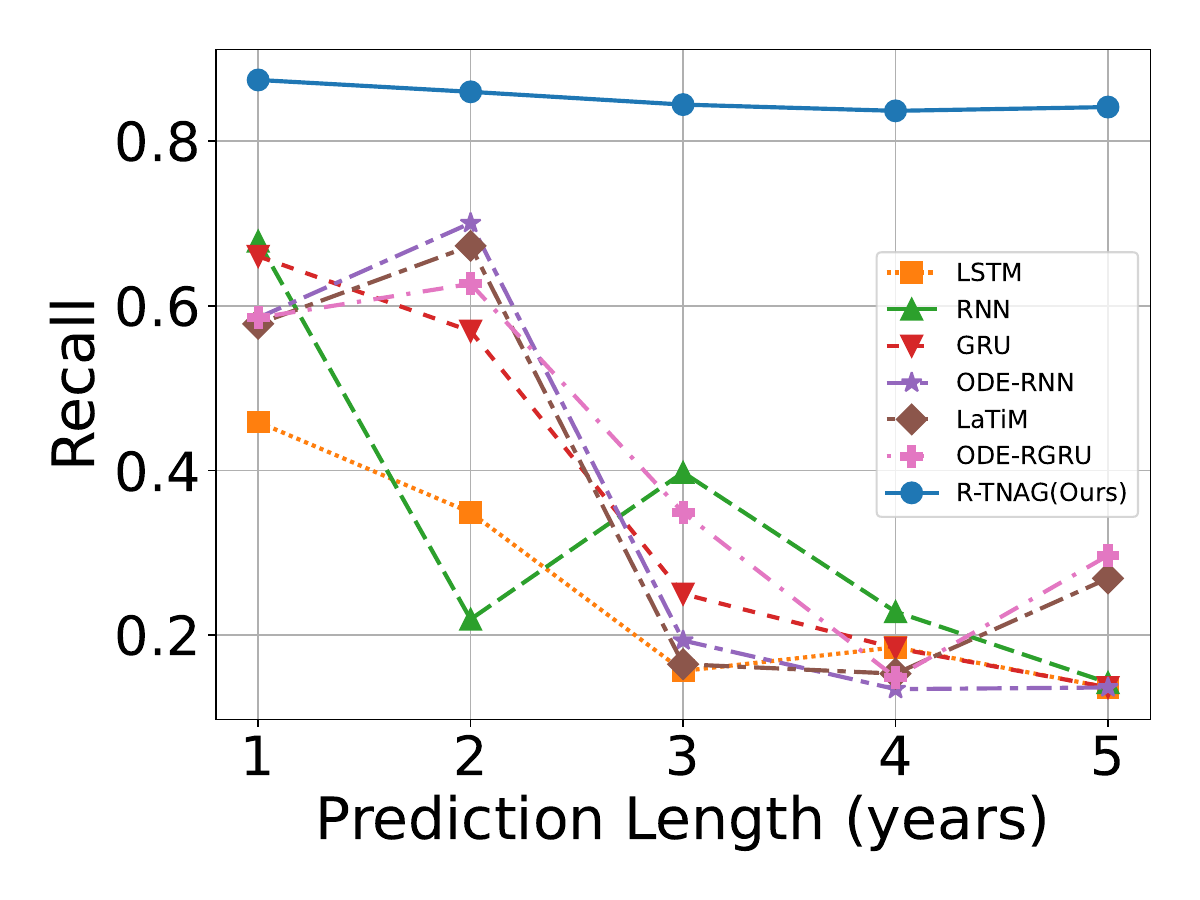}}
    \hfill
    \subfloat{\includegraphics[width=0.24\linewidth]{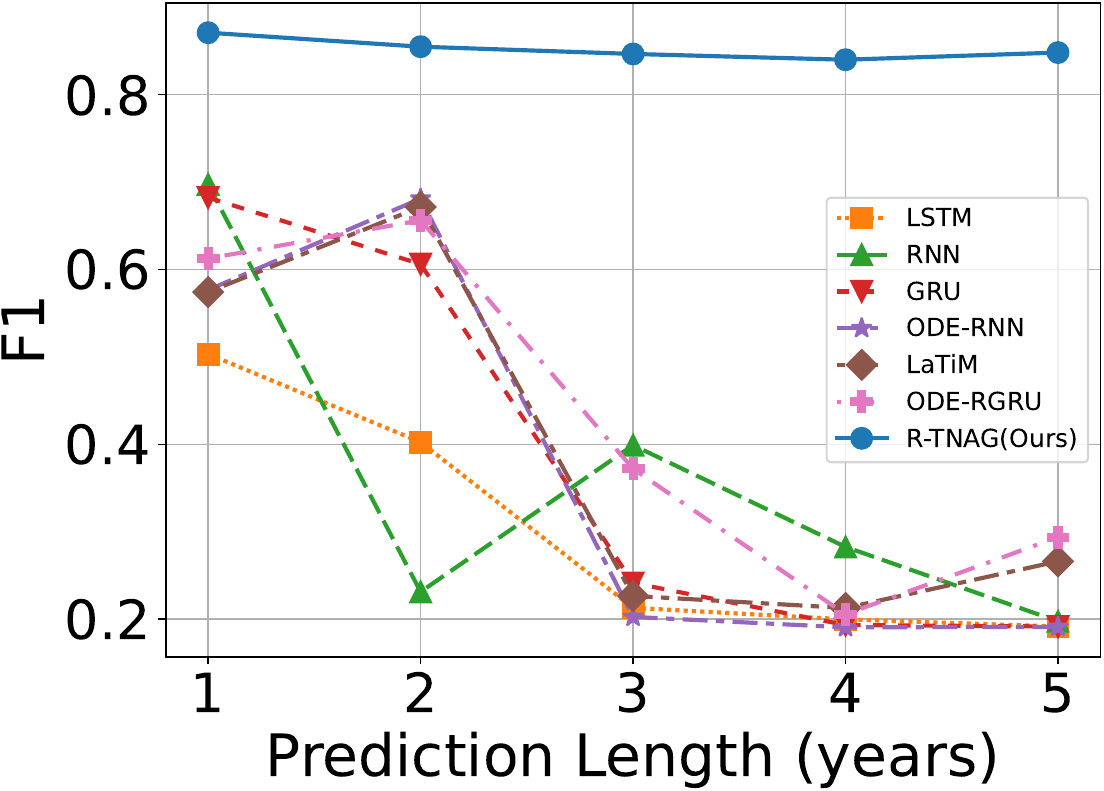}}
\centering
    \caption{Longitudinal Evaluation of Disease State Prediction over 1–5 Years.}
    
\label{fig:classification_accuracy}
\end{figure*}

To assess R-TNAG’s effectiveness in irregular-interval prediction, we compared it with conventional temporal models and continuous-time models in Euclidean and manifold spaces. Using 1–5 year observation sequences as inputs, we evaluated disease state prediction (CN, MCI, AD). Results are shown in \cref{fig:classification_accuracy}.
The comparative models encompass representative baselines, including RNN \cite{jung2021deep}, LSTM \cite{rnn_lstm_gru2019}, and GRU \cite{park2024predicting} for discrete-time modeling in Euclidean space; ODE-RNN \cite{rubanova2019latent} and LaTiM \cite{LaTiM} for continuous-time modeling in Euclidean space; and ODE-RGRU \cite{odergru_ad} for manifold-based NODE approaches. These baselines provide a comprehensive benchmark to highlight the advantages of the proposed method in manifold modeling and temporal feature representation.

As shown in \cref{fig:classification_accuracy}, extending the prediction horizon from 3 to 5 years causes a marked decline in mAUC, Precision, Recall, and F1 across all methods, highlighting the challenge of long-term forecasting with irregularly sampled data.
Continuous-time models (ODE-RNN, LaTiM, ODE-RGRU) drop sharply from years 3 to 4 but show a modest rebound at year 5, suggesting that continuous representations in both Euclidean and manifold spaces are better suited for handling irregular follow-up intervals and long-range extrapolation. In particular, ODE-RNN shows a slight upward trend in mAUC and Precision from year 4 onward. LSTM exhibits a similar recovery in mAUC at years 4–5 and in Precision at year 5, indicating improved reliability in identifying true AD cases and a retained capacity to model long-term dependencies. In contrast, RNN and GRU display a monotonic decline, with RNN showing particularly unstable behavior—sharp drops at year 2, a small rebound at year 3, followed by continued deterioration through years 4 and 5.

Notably, R-TNAG sustains consistently high accuracy and stability across all 1–5-year predictions. These findings underscore the advantages of continuous-time modeling for capturing long-term disease dynamics and managing irregular intervals, highlighting the robustness and practical value of R-TNAG for irregular-interval prediction tasks.

\subsection{Comparative Analysis of 5-Year Cognitive Regression}

To evaluate R-TNAG’s regression performance in predicting cognitive scores, we used 5-year longitudinal data to forecast the final time-step MMSE, ADAS11, and ADAS13 scores (\cref{table:Comparison2}).

Discrete-time Euclidean models (RNN, LSTM, GRU) produced large errors and negative $R^2$ values; for example, GRU yielded $R^2 = -21.88$ on MMSE, reflecting poor capability in long-term irregular prediction. Continuous-time Euclidean models (ODE-RNN, LaTiM) improved stability but still struggled with consistent trend fitting. The manifold-based ODE-RGRU achieved positive $R^2$ with reduced errors; however, interpolation introduced deviations in trajectory reconstruction, leading to suboptimal trend fitting.

R-TNAG achieved the lowest errors for MMSE and ADAS11 and the highest $R^2$ across all three scores, demonstrating both accurate prediction and robust trend fitting. These findings indicate that R-TNAG effectively captures the continuity of cognitive decline, balancing local accuracy with global trajectory preservation.

\begin{table*}[!t]
\centering
\caption{Comparative Analysis of 5-Year Cognitive Regression}
\label{table:Comparison2}
\resizebox*{\textwidth}{!}{

    \begin{tabular}{lcccccc}
        \toprule
        \multirow{2}{*}{Method} & \multicolumn{2}{c}{MMSE} & \multicolumn{2}{c}{ADAS11} & \multicolumn{2}{c}{ADAS13} \\
        \cmidrule(lr){2-3} \cmidrule(lr){4-5} \cmidrule(lr){6-7}
        & MAPE ↓ & R$^2$ ↑ & MAPE ↓ & R$^2$ ↑ & MAPE ↓ & R$^2$ ↑ \\
        \midrule
RNN\cite{jung2021deep} & 0.1038 ± 0.0428          & 0.0167 ± 0.5548          & 0.6745 ± 0.2251          & 0.2394 ± 0.3411          & 0.6152 ± 0.2415          & 0.1895 ± 0.3977 
 \\
LSTM\cite{rnn_lstm_gru2019} & 0.2660 ± 0.2020            & -6.2168 ± 7.7744         & 0.8530 ± 0.1085           & -0.4015 ± 0.4530          & 0.8180 ± 0.1427           & -0.2503 ± 0.2028
 \\
GRU\cite{park2024predicting} & 0.4840 ± 0.3047           & -21.8819 ± 21.6558       & 1.0027 ± 0.4552          & -1.7159 ± 2.5334         & 0.9429 ± 0.2825          & -1.3675 ± 1.7085
 \\
ODE-RNN\cite{rubanova2019latent} & 0.2136 ± 0.2523          & -6.3925 ± 13.3537        & 0.5805 ± 0.1847          & 0.3260 ± 0.3490            & 0.5252 ± 0.0814          & 0.2490 ± 0.5488
\\
LaTiM\cite{LaTiM} & 0.1869 ± 0.2415          & -3.9825 ± 9.0293         & 0.6878 ± 0.5890           & -0.1615 ± 1.5925         & \textbf{0.4491 ± 0.1803} & 0.2638 ± 0.8706 \\
ODE-RGRU\cite{odergru_ad} & 0.0844 ± 0.0184          & 0.3433 ± 0.1781          & 0.6750 ± 0.3688           & 0.3028 ± 0.2076          & 0.5943 ± 0.2448          & 0.3668 ± 0.1574
 \\
R-TNAG(Ours) & \textbf{0.0811 ± 0.0174} & \textbf{0.3751 ± 0.0913} & \textbf{0.5412 ± 0.0791} & \textbf{0.3609 ± 0.1087} & 0.5431 ± 0.0861          & \textbf{0.4135 ± 0.1033}
 \\ \bottomrule  
 
\end{tabular}
 }
\end{table*}

\subsection{Model Comparison Across Missing Data Rates}

To assess model performance on irregularly sampled time series, we conducted disease state prediction and cognitive score regression using 5-year longitudinal data. The original dataset had a 39.2\% missing rate, to which we added 0.1, 0.3, and 0.5 additional missingness, yielding overall rates of about 49.2\%, 69.2\%, and 89.2\%. These settings simulate progressively sparser clinical follow-up scenarios, where observations may be missing due to variability, incomplete data collection, or equipment failure. Performance was evaluated using mAUC for disease prediction and MAPE for score regression.

\begin{table*}
\centering
\caption{Prediction Performance at Different Missing Rates}
\label{table:Comparison5}
\resizebox*{0.87\textwidth}{!}{
\begin{tabular}{l|c|cc|c|cc} \toprule
Method     & Missing Rate  & mAUC$\uparrow$                      & MAPE$\downarrow$       & Missing Rate          & mAUC$\uparrow$                      & MAPE$\downarrow$ \\ \midrule
RNN\cite{jung2021deep}       & \multirow{7}{*}{39.2(0)}     & 0.6781 ± 0.1010  & 0.4645 ± 0.1698         & \multirow{7}{*}{49.2(0.1)} & 0.6477 ± 0.0997  & 0.5913 ± 0.1124         \\
LSTM\cite{rnn_lstm_gru2019}      &                      & 0.6017 ± 0.0387  & 0.6457 ± 0.1511  & ~ & 0.5636 ± 0.0605  & 0.6363 ± 0.1897\\
GRU\cite{park2024predicting}       &                      & 0.6337 ± 0.0632  & 0.8099 ± 0.3475  & ~ & 0.7166 ± 0.0577  & 1.0377 ± 0.3591\\
ODE-RNN\cite{rubanova2019latent}  &                      & 0.7052 ± 0.1047  & 0.4398 ± 0.1728  & ~ & 0.7052 ± 0.1047  & 0.4398 ± 0.1728\\
LaTiM\cite{LaTiM}     &                      & 0.7079 ± 0.0853  & 0.4413 ± 0.3369  & ~ & 0.7079 ± 0.0853  & 0.4413 ± 0.3369\\
ODE-RGRU\cite{odergru_ad} &                      & 0.6835 ± 0.0425  & 0.4512 ± 0.2107  & ~ & 0.6602 ± 0.0558 & 0.4307 ± 0.1384\\
R-TNAG(Ours)&                      & \textbf{0.9582 ± 0.0224}& \textbf{0.3885 ± 0.0609}  & ~ & \textbf{0.9514 ± 0.0291} & \textbf{0.4295 ± 0.0630}\\ \midrule
RNN\cite{jung2021deep}       & \multirow{7}{*}{69.2(0.3)} & 0.5824 ± 0.0668  & 0.5826 ± 0.1176          & \multirow{7}{*}{89.2(0.5)} & 0.6022 ± 0.0422  & 0.6663 ± 0.2275\\
LSTM\cite{rnn_lstm_gru2019}      &                      & 0.5332 ± 0.0556  & 0.5948 ± 0.1220  & ~ & 0.5351 ± 0.0537  & 0.6047 ± 0.1897\\
GRU\cite{park2024predicting}       &                      & 0.7184 ± 0.0444  & 0.8111 ± 0.3828  & ~ & 0.6637 ± 0.0393  & 0.9552 ± 0.2796\\
ODE-RNN\cite{rubanova2019latent}  &                      & 0.6849 ± 0.0826  & 0.4280 ± 0.1732  & ~ & 0.6497 ± 0.0725  & 0.4124 ± 0.1968\\
LaTiM\cite{LaTiM}     &                      & 0.7100 ± 0.0957  & 0.4415 ± 0.3388  & ~ & 0.7316 ± 0.0585  & 0.4485 ± 0.3426 \\
ODE-RGRU\cite{odergru_ad} &                      & 0.6437 ± 0.0394  & 0.4910 ± 0.1545 & ~ & 0.7372 ± 0.0621 & 0.4117 ± 0.0737\\
R-TNAG(Ours) &                      & \textbf{0.9432 ± 0.0327}  & \textbf{0.4229 ± 0.0803}  & ~ & \textbf{0.9268 ± 0.0368}  & \textbf{0.3923 ± 0.0506}\\ \bottomrule
\end{tabular}
}
\end{table*}

As shown in \cref{table:Comparison5}, discrete-time Euclidean models (RNN, LSTM, GRU) are highly sensitive to missing data, with mAUC dropping sharply and MAPE rising even at low sparsity—for example, GRU’s MAPE exceeded 1.0 at a 49.2\% missing rate. Continuous-time models (ODE-RNN, LaTiM) perform more stably with lower errors but still degrade as missingness increases. The manifold-based ODE-RGRU further reduces errors and maintains positive mAUC, yet shows noticeable prediction bias and weaker trend fitting under high sparsity.
By contrast, R-TNAG consistently achieves the best performance, with mAUC of 0.9268 and MAPE of 0.3923 even at 89.2\% missingness, demonstrating robust disease classification and cognitive score regression under irregular and incomplete observations. These results confirm R-TNAG’s strength in preserving predictive accuracy and trajectory consistency, making it well suited for modeling long-term disease progression in sparse or unevenly sampled clinical data.

\subsection{Cross-Dataset Evaluation of Disease Prediction}

To assess robustness across datasets, 5-year longitudinal data from ADNI1 and ADNI2 were used for disease state prediction. Five-fold cross-validation was performed on ADNI1, with results reported as mean ± standard deviation, while ADNI2 served as an independent external test set to evaluate generalization. The results are shown in \cref{table:Comparison3}.

Discrete-time Euclidean models (RNN, LSTM, GRU) exhibit limited accuracy and stability, with low mAUC and F1 on both ADNI1 and ADNI2. Continuous-time models (ODE-RNN, LaTiM) improve performance, but F1 scores on ADNI2 remain modest, indicating challenges in cross-dataset generalization. The manifold-based ODE-RGRU further increases mAUC and F1, particularly on ADNI2, but shows inconsistent trend fitting, with ADNI1 achieving only 0.8401 mAUC and 0.6181 F1.

In contrast, R-TNAG consistently attains high mAUC and F1 on ADNI1 (0.9513 and 0.8375, respectively) and competitive results on ADNI2, demonstrating strong robustness, stable classification, and practical utility under dataset shifts.

\begin{table}
\centering
\caption{Cross-Dataset Comparison of Disease State Prediction }
\label{table:Comparison3}
\resizebox{\columnwidth}{!}{
\begin{tabular}{lcc|cc}
\toprule
\multirow{2}{*}{Method} & \multicolumn{2}{c|}{ADNI 1} & \multicolumn{2}{c}{ADNI 2} \\
\cmidrule(lr){2-3} \cmidrule(lr){4-5}
 & mAUC$\uparrow$ &F1$\uparrow$ &mAUC$\uparrow$ &F1$\uparrow$ \\
\midrule
RNN\cite{jung2021deep} &  0.7287 ± 0.1272 & 0.2983 ± 0.1743 & 0.7343 & 0.5070 \\
 LSTM\cite{rnn_lstm_gru2019} & 0.6698 ± 0.0892 & 0.1578 ± 0.0211 & 0.5687 & 0.2639 \\
GRU\cite{park2024predicting}    &0.7498 ± 0.0915          & 0.1665 ± 0.0291          & 0.6780          & 0.0808 \\
ODE-RNN \cite{rubanova2019latent} & 0.7910 ± 0.0548          & 0.1785 ± 0.0441          & 0.7405          & 0.4955 \\
LaTiM \cite{LaTiM} & 0.8424 ± 0.0489          & 0.3093 ± 0.1596          & 0.7500          & 0.2639  \\
ODE-RGRU \cite{odergru_ad} & 0.8401 ± 0.0745          & 0.6181 ± 0.1993          & \textbf{0.8787} & 0.6254 \\
R-TNAG(Ours)  & \textbf{0.9513 ± 0.0352} & \textbf{0.8375 ± 0.0685} & 0.8479          & \textbf{0.6718} \\
\bottomrule
\end{tabular}
}
\end{table}

\section{Conclusion} 
\label{sec:concluion}

This study proposes R-TNAG for disease progression prediction from irregular-interval time series, combining 3D sMRI manifold mapping with time-aware continuous modeling. Its three modules: RMM, TNODE, and ARGRU, synergistically extract spatial features, model continuous temporal dynamics, and handle irregular intervals, enabling accurate 5-year predictions of disease states and cognitive trajectories.

Experimental results show that R-TNAG effectively captures long-term disease dynamics and handles irregular follow-up intervals in 1–5-year longitudinal disease state prediction, highlighting its robustness and practical utility. In the 5-year cognitive score regression, the model preserves the continuity of cognitive decline, balancing local accuracy with global trajectory trends. Across varying missing data rates, R-TNAG consistently maintains high predictive accuracy and trajectory consistency, confirming its suitability for sparse or unevenly sampled clinical data. In cross-dataset evaluation with ADNI1 and ADNI2, it achieves a high mAUC and F1 on ADNI1 and demonstrates competitive performance on ADNI2, indicating stable classification and strong generalization.

In summary, R-TNAG effectively captures temporal imaging features and delivers reliable predictions from irregular-interval data, providing a valuable framework for future longitudinal disease modeling and related clinical applications.

\bibliographystyle{IEEEtran}
\bibliography{ref.bib}

\end{document}